\documentclass{article}

\usepackage[preprint]{neurips_2026}

\usepackage[utf8]{inputenc}
\usepackage[T1]{fontenc}
\usepackage[hypertexnames=false]{hyperref}
\usepackage{url}
\usepackage{booktabs}
\usepackage{amsfonts,amssymb}
\usepackage{nicefrac}
\usepackage{microtype}
\usepackage{xcolor}
\usepackage{graphicx}
\usepackage{amsmath}
\usepackage{amsthm}
\usepackage{float}

\newtheorem{theorem}{Theorem}
\newtheorem{assumption}{Assumption}
\newtheorem{corollary}{Corollary}

\title{Learning Under Forgetting: Statistical Support-Selective Retention in Stochastic Training Dynamics}

\author{Fujie Gao \And Zuyue Zhang \And Gang Sun}

\hypersetup{hidelinks, pdftitle={Learning Under Forgetting: Statistical Support-Selective Retention in Stochastic Training Dynamics}, pdfauthor={Fujie Gao, Zuyue Zhang, Gang Sun}}

\begin{document}

\maketitle

\begin{abstract}
Prior work has shown that neural networks exhibit implicit biases toward low-complexity structure (e.g., spectral bias), memorization dynamics, and compression-like effects during training, but a unified dynamical account of selective retention remains incomplete. We propose Repeated Reinforcement with Persistent Forgetting (RPF) dynamics, a minimal framework in which repeated exposure reinforces patterns and structures that recur in the data, while persistent forgetting attenuates learned information. This view treats forgetting not merely as a failure mode, but as a selection mechanism. We build the theory in three successive layers. First, in an independent-feature model, we derive an exposure-selective survival law and a support-dependent retention boundary characterizing which patterns persist under forgetting. Second, in a shared-parameter model, we show that forgetting induces spectral filtering over covariance modes, preserving strongly supported shared components while suppressing weak ones. Third, under small-step and norm/coding approximations, we show how RPF dynamics induce an implicit trade-off between data fitting and the cost of stored information, yielding Minimum Description Length (MDL)-like compression. Controlled experiments provide evidence for this reinforcement--forgetting selection mechanism in scalar memories and a nonlinear shared network. Joint reinforcement and attenuation interventions shift conditional retention, while matched exposure counts reveal forgetting-dependent effects of reinforcement timing and changes in the composition of the retained set. Together, these results show that repeated reinforcement and persistent forgetting jointly provide a controllable source of inductive bias beyond neural architecture and scale.
\end{abstract}

\section{Introduction}

Training is usually described as a process of accumulating information from data, while mechanisms such as weight decay and $\ell_2$ regularization \citep{krogh1992simple,loshchilov2019decoupled}, dropout \citep{Srivastava2014}, and stochastic updates \citep{mandt2017sgd} are treated as regularization or noisy optimization dynamics. From the perspective of training dynamics, however, these mechanisms can also attenuate, mask, perturb, or overwrite learned representations. This shifts the question from how to prevent forgetting to what forgetting permits to survive.

This is a fundamental issue for modern learning systems trained on large, redundant data streams. Such systems are exposed to vast amounts of raw information, yet often retain stable patterns and structures while discarding incidental details. A sharper version arises when the available experience is limited but highly redundant: a system may still acquire reusable patterns and structures without retaining broad factual detail. Data do not support all information equally. Patterns and structures that recur in the data receive repeated reinforcement through many updates, whereas weakly supported details receive inconsistent reinforcement. If learning and forgetting coexist throughout training, which patterns and structures are retained, which are forgotten, and what determines this separation?

Prior work has studied forgetting, regularization, memorization, and scaling, but has paid relatively little attention to directly characterizing which patterns survive under simultaneous reinforcement and attenuation.

We address this selective-retention problem from the perspective of training dynamics. Throughout this paper, patterns and structures denote reusable units of regularity, such as words or relations in language, edges or textures in images, timbre or recurring acoustic motifs in sound, and shared directions in learned representations. Repeated Reinforcement with Persistent Forgetting (RPF) is not a semantic importance criterion: frequent superficial patterns may survive, while rare meaningful facts may require additional mechanisms. Its central answer is that information is retained when it can be reconstructed often enough to survive forgetting. Our contributions are:

\begin{itemize}
\item We formulate RPF dynamics to study training-dynamics-induced selective retention under repeated exposure and persistent forgetting;
\item We derive the survival mechanism from an independent-feature exposure law to shared-parameter spectral filtering, and give an approximate MDL-like interpretation;
\item We test this mechanism in scalar memories and in a shared nonlinear network, where a reinforcement--forgetting phase diagram and matched occurrence counts reveal conditional and history-dependent selection.
\end{itemize}

\section{Related Work}
\label{sec:related-work}

We relate RPF to four lines of work: forgetting, training dynamics, data repetition, and compression. Forgetting has been studied most extensively in continual learning, where it is usually treated as a failure mode to be mitigated. Foundational work identified catastrophic interference as a central limitation of connectionist systems \citep{mccloskey1989catastrophic,french1999catastrophic,goodfellow2013empirical}. Subsequent methods preserve knowledge through explicit constraints, importance estimates, distillation, replay, or memory modules, including EWC and Synaptic Intelligence \citep{kirkpatrick2017overcoming,zenke2017continual}, Learning without Forgetting \citep{li2016learning}, experience replay \citep{rolnick2019experience}, GEM and A-GEM \citep{lopez2017gradient,chaudhry2019efficient}, iCaRL \citep{rebuffi2017icarl}, and memory architectures \citep{weston2014memory}. Recent work extends these concerns to large-model lifelong learning and stability--plasticity trade-offs \citep{ven2024continual,dasbiswas2026ella,lu2025rethinking}. These approaches ask how to prevent or compensate for forgetting; RPF asks which repeatedly reinforced patterns and structures survive when forgetting remains active.

RPF also relates to learning dynamics and regularization. Gradient descent can converge to particular solutions in underdetermined settings \citep{soudry2018implicit,gunasekar2018implicit}, neural networks exhibit spectral bias \citep{rahaman2019spectral}, and example-level training dynamics reveal that some examples are repeatedly forgotten while others are never forgotten \citep{toneva2019example}. Grokking shows delayed transitions from memorization to generalization \citep{power2022grokking}, with later work emphasizing weight decay, data-size dependence, and representation emergence \citep{liu2023omnigrok}. Weight decay and dropout are usually viewed as regularizers \citep{krogh1992simple,loshchilov2019decoupled,Srivastava2014}, and stochastic optimization can be studied as noise-induced dynamics \citep{mandt2017sgd}. Our focus is not regularization as such, but whether a mechanism implements a forgetting pressure that competes with pattern-specific reinforcement; in our experiments, weight decay provides a direct smooth-attenuation knob, while dropout is treated as a candidate masking-based mechanism whose role must be evaluated empirically.

More directly, \citet{yunis2024spectraldynamics} relate weight singular values, effective rank, and alignment to generalization, and show that weight decay strengthens rank bias beyond norm control. \citet{lu2025frequency} analyze Fourier-frequency suppression in convolutional weights under regularization. RPF uses covariance-supported directions in its shared-linear case and makes pattern-level retention boundaries and reinforcement schedules explicit intervention objects. \citet{balasubramanian2026posttraining} quantify old-mode mass loss and component drift through divergence direction, behavioral overlap, sampling, and replay in a two-mode post-training model. RPF examines selection across nonuniformly reinforced patterns under persistent attenuation, including how matched occurrence counts acquire different final value through their timing. The contribution is this retention formulation and its testable intervention structure.

Selective retention is also an empirical concern in large language models. LLMs can emit verbatim training data \citep{carlini2021extracting}, and memorization increases with model size, context, and duplicated examples \citep{carlini2023quantifying,biderman2023emergent}. Training-dynamics studies show that larger models memorize faster and forget less \citep{tirumala2022memorization}; data-constrained scaling work shows that repeated data can help up to a point \citep{muennighoff2023scaling}; long-tail knowledge studies show that factual recall depends strongly on pretraining evidence \citep{kandpal2023longtail}; and controlled knowledge-storage work shows that data diversity and augmentation affect whether stored knowledge becomes extractable \citep{allenzhu2024physics}. These findings motivate a mechanistic question: how does repeated exposure translate into retention under persistent forgetting?

Finally, scaling laws characterize aggregate performance as a function of model size, data, and compute \citep{kaplan2020scaling,hoffmann2022training,henighan2020scaling,paquette2024phases}; related quantization views model scaling as the acquisition of discrete skills or quanta in frequency order \citep{michaud2024quantization}. Natural data and learned examples are highly nonuniform, often exhibiting Zipf-like, power-law, or long-tailed structure \citep{zipf1949human,newman2005power,piantadosi2014zipf,simoncelli2001natural,zhang2024longtailed}, with long-tail memorization playing an important role in neural networks \citep{feldman2020memorize}. RPF connects these views at the level of retained information rather than only loss. RPF also differs from explicit Information Bottleneck or MDL objectives \citep{tishby2015deep,alemi2017deep,kawaguchi2023information,rissanen1978modeling,grunwald2007minimum,hinton1993keeping}: the compression-like effect arises from the training dynamics, through attenuation of components that are not repeatedly reconstructed.

\section{Theoretical Analysis}

\subsection{Formalization}

\paragraph{Patterns and structures.}
Let $\mathcal{S}$ denote a space of reusable regularities in the data stream. A member $s\in\mathcal{S}$ is a pattern whose occurrences induce compatible learning signals. The boundary of $s$ is operational rather than semantic: it may be a word, phrase, relation, texture, acoustic motif, or a distributed component that can be repeatedly encountered, reinforced, weakened, and reconstructed during training.

\paragraph{Persistent forgetting.}
Persistent forgetting continually reduces the recoverable strength of previously encoded information. In RPF, forgetting is a conceptual pressure on stored patterns and structures; concrete training techniques such as decay, regularization, stochastic perturbation, dropout-style masking, overwriting, or interference may implement this pressure in different ways. Whether a particular technique realizes the desired pattern-level forgetting effect is setting-dependent. Let $\gamma_t(s)\ge0$ denote the effective forgetting pressure on pattern $s$ at step $t$. We analyze multiplicative decay with rate $d$ as the tractable smooth-attenuation case, for which a typical forgetting strength satisfies $\gamma\approx d$ up to representation- and optimizer-dependent constants.

\paragraph{Exposure and repeated reinforcement.}
Let $p(s)$ be the data exposure of $s$. Repeated traversal increases the number of steps $n$, while selective replay or sampling changes may alter the sampling exposure $q(s)$; otherwise $q(s)=p(s)$. Loss reweighting instead changes the compatible update per exposure, represented by $R(s)$ under a fixed learning-strength scale. For $n$ training steps, let $E_t(s)$ be the fraction of examples in step $t$'s minibatch that expose $s$ (an indicator for a single-example step), and define normalized cumulative exposure
\begin{equation}
N_n(s)=\sum_{t=1}^n E_t(s), \qquad \mathbb{E}[N_n(s)] = n q(s)
\end{equation}
when minibatch examples are sampled independently from $q$. Thus repeated exposure creates repeated reinforcement opportunities.

Repeated reinforcement is accumulated compatible update pressure, not raw frequency alone. Let $\rho_t(s)\ge0$ denote the compatible reinforcement coefficient at step $t$, normalized by the remaining recovery gap and measured before subsequent attenuation. Frequent and consistently aligned occurrences make $\sum_t\rho_t(s)$ grow, whereas rare or inconsistent occurrences yield weak accumulated reinforcement. For discrete patterns, $q(s)$ may be estimated by exposure frequency; for distributed structures, effective support may arise through covariance strength, shared contexts, replay, weighting, or other mechanisms that repeatedly align updates.

\paragraph{Survival and recoverability.}
A pattern survives if it remains usable after repeated weakening and relearning. Let $R(s)\ge0$ denote its local recovery coefficient per compatible exposure, normalized by the learning-strength scale $\eta$ and the remaining gap to a reference strength. Exposure frequency is accounted for separately by $q(s)$: $R(s)$ measures recovery given an exposure, rather than inverse recovery time measured in ordinary training steps. Recoverability depends on the current representation and update protocol. Exposure and this conditional recovery coefficient combine into the effective support
\begin{equation}
\lambda_{\mathrm{eff}}(s)=q(s)R(s).
\end{equation}
Low-support details have small $q(s)$; unstable structures have small $R(s)$; both are vulnerable to persistent forgetting. The analytic models below instantiate this support scale in two basis cases: the independent-feature model has $R(s)=1$ so that support reduces to exposure, while the shared-parameter model replaces isolated exposure by covariance-supported directions. Section~\ref{sec:discussion} specifies the normalization and distinguishes conditional recovery from recovery time in the training stream.

\subsection{Repeated Reinforcement with Persistent Forgetting (RPF)}

RPF couples reinforcement and forgetting in one training dynamics. Let $Z_t$ denote the learned state at step $t$, abstracting over parameters, representations, or memory traces. RPF writes one coarse-grained training step as an update operator followed by a forgetting operator:
\begin{equation}
Z_{t+1}=\mathcal{F}_t\bigl(\mathcal{U}_t(Z_t;\mathcal{B}_t)\bigr),
\label{eq:rpf-general}
\end{equation}
where $\mathcal{B}_t$ is the minibatch or data event at step $t$. The update $\mathcal{U}_t$ reinforces data-supported structures, while $\mathcal{F}_t$ attenuates learned information. Thus retained structures must be repeatedly reconstructed against persistent attenuation.

Under stationary local mean-field bookkeeping, the nominal reinforcement scale is
\begin{equation}
\mathbb{E}\!\left[\sum_{t=1}^n \rho_t(s)\right]
\propto n\eta\,\lambda_{\mathrm{eff}}(s)
=n\eta\, q(s) R(s),
\label{eq:cumulative-reinforcement}
\end{equation}
where $\eta$ is a learning-strength scale. Final retained strength also depends on the survival of these updates through subsequent dynamics. With effective forgetting strength $\gamma$, a schematic finite-horizon reconstruction requirement takes the form
\begin{equation}
n\eta\,\lambda_{\mathrm{eff}}(s) \ge C_n(\gamma,\epsilon).
\label{eq:finite-survival}
\end{equation}
Here $C_n$ depends on the horizon as well as attenuation and the retention criterion; it cannot be held fixed while persistent forgetting accumulates. The scalar case below supplies an exact finite-horizon law. In steady state, the characteristic competition is between the per-step scales,
\begin{equation}
\eta\,\lambda_{\mathrm{eff}}(s) \gtrsim \gamma.
\label{eq:steady-survival}
\end{equation}
Thus patterns and structures persist when their effective support is large enough to withstand persistent forgetting. The finite-time condition above concerns cumulative reconstruction opportunities, while the steady-state condition compares per-step reinforcement with per-step attenuation. In the independent-feature model below, $\lambda_{\mathrm{eff}}$ reduces to exposure up to the learning scale; in the shared-parameter model, the analogous support quantity is a covariance eigenvalue. The specified retention level determines the boundary constant, and the mapping to effective support is representation-dependent; Appendix~\ref{app:theory-details} gives additional bookkeeping details.

\subsection{Theoretical Results}

We develop this reinforcement--forgetting selection in three successive layers: survival of independent patterns, retention of shared representation directions, and an information-constrained interpretation of maintained structure.

\subsubsection{Independent features: exposure-selective survival}

We first isolate the mechanism by assigning each pattern an independent scalar strength.

Let $x_s^t\in\{0,1\}$ indicate whether pattern $s$ is exposed at step $t$, let $w_s^t\in\mathbb{R}$ denote its learned strength, and let $\theta_s$ be the fully learned target value. We use the multiplicative-forgetting dynamics
\begin{equation}
w_s^{t+1}=(1-d)\left[w_s^t+\eta x_s^t(\theta_s-w_s^t)\right],
\label{eq:independent-dynamics}
\end{equation}
where $d\in(0,1)$ is the forgetting rate and $\eta>0$ is the learning rate.

\begin{assumption}[Independent-feature RPF]
\label{assump:independent-rpf}
For each $s$, exposure events are sampled independently over time with $x_s^t\sim\mathrm{Bernoulli}(p(s))$. Updates are compatible with the target $\theta_s$, forgetting follows Eq.~\eqref{eq:independent-dynamics} with $0<d<1$, $0<\eta p(s)<1$, and $|\theta_s|<\infty$.
\end{assumption}

\begin{theorem}[Exposure-selective survival]
\label{thm:exposure-selective}
Under Assumption~\ref{assump:independent-rpf}, $\mathbb{E}[w_s^t]$ converges to
\begin{equation}
w_s^*=\frac{(1-d)\eta p(s)}{d+(1-d)\eta p(s)}\,\theta_s.
\label{eq:independent-fixed-point}
\end{equation}
Consequently there is a characteristic survival scale
\begin{equation}
p^*=\Theta\!\left(\frac{d}{(1-d)\eta}\right)
\label{eq:p-threshold}
\end{equation}
such that
\begin{equation}
w_s^*\approx
\begin{cases}
\dfrac{(1-d)\eta p(s)}{d}\theta_s, & p(s)\ll p^*,\\[6pt]
\theta_s, & p(s)\gg p^*.
\end{cases}
\end{equation}
\end{theorem}

Thus retention is controlled by the ratio between reinforcement $(1-d)\eta p(s)$ and forgetting $d$, producing a soft exposure threshold. For $\theta_s\ne0$ and $0<\epsilon<1$, the criterion $w_s^*/\theta_s\ge\epsilon$ is equivalent to
\begin{equation}
p(s)\ge p_\epsilon^*=\frac{\epsilon}{1-\epsilon}\frac{d}{(1-d)\eta}.
\label{eq:epsilon-boundary}
\end{equation}
This is a selection rule for the retained set at the specified strength level. The theorem characterizes expected strength; finite-run retention remains stochastic. The proof and finite-run variability details are in Appendix~\ref{app:proofs} and Appendix~\ref{app:theory-details}.

\paragraph{Heavy-tailed exposure and finite-time scaling.}
Under heavy-tailed exposure, the independent-feature law yields a retained-count scaling.

\begin{assumption}[Power-law exposure]
\label{assump:power-law}
Let patterns $s_r$ obey a power-law exposure profile,
\begin{equation}
p(s_r)=C r^{-\alpha},\qquad \alpha>1,
\end{equation}
where $r=1,2,\ldots$ is the decreasing-exposure rank, $C>0$ is the normalizer, and $\alpha$ is the tail exponent.
\end{assumption}

The exact power-law form is used only to obtain a closed-form retained-count scaling. The RPF survival boundary itself requires only nonuniform effective support with scale separation.

For a finite run of length $n$, define the effective training strength
\begin{equation}
S=n\eta(1-d).
\label{eq:effective-strength}
\end{equation}
In the accumulation-dominated regime, where $nd\ll1$ and the discrete learning step is small at the boundary, a fixed criterion on normalized expected strength gives $S p(s_r)=\Theta(1)$. Section~\ref{sec:discussion} derives this regime and its crossover to the forgetting-limited equilibrium.
\begin{corollary}[Finite-time retained-count scaling]
\label{cor:finite-scaling}
Under Assumption~\ref{assump:power-law} in the accumulation-dominated regime with $S p(s_r)=\Theta(1)$, the retained rank cutoff $k^*$, defined by $S p(s_{k^*})=\Theta(1)$, scales as
\begin{equation}
k^*\sim S^{1/\alpha}
\label{eq:finite-scaling}
\end{equation}
up to constants depending on $C$ and $\epsilon$, while the accumulation-dominated approximation holds and before exhausting the pattern support.
\end{corollary}
Thus early training effort expands the expected-strength retained set sublinearly. At fixed positive forgetting, further training approaches a retention ceiling rather than extending this power law indefinitely. Under heavy-tailed exposure, this also means that a relatively small retained set can cover a large fraction of exposure mass. The derivation and retained-mass estimate are in Appendix~\ref{app:theory-details}.

\subsubsection{Shared parameters: spectral filtering}

We next show that when parameters are shared, the same reinforcement--forgetting balance selects supported directions of variation rather than isolated patterns. This is the key shift from retaining independent pattern coordinates to retaining reusable directions: in shared representations, forgetting no longer asks whether one pattern survives alone, but which directions are rebuilt by many compatible patterns and structures.

Consider a linear prediction setting in which a sampled pattern $s$ is represented by a feature vector $x=x(s)\in\mathbb{R}^m$, with target parameter $\theta\in\mathbb{R}^m$ and model parameter $w\in\mathbb{R}^m$. Let
\begin{equation}
\Sigma=\mathbb{E}[xx^\top]
\end{equation}
be the second-moment matrix of pattern representations (the covariance for centered features). For discrete pattern inputs, $\Sigma=\sum_s p(s)x(s)x(s)^\top$ is exposure-weighted. Its eigenspectrum measures support along shared directions.

\begin{assumption}[Shared-parameter RPF]
\label{assump:shared-rpf}
The expected update is governed by the second-moment matrix $\Sigma$ above, forgetting applies the same multiplicative attenuation rate $d$ to all coordinates after each expected update, and $0<\eta\lambda_i<1$ for every nonzero mode.
\end{assumption}

For the linear predictor $x^\top w$ with target $x^\top\theta$ and squared loss, the population gradient is $\Sigma(w-\theta)$. Applying multiplicative attenuation after the gradient update gives the exact expected dynamics
\begin{equation}
w^{t+1}=(1-d)\left[w^t+\eta\Sigma(\theta-w^t)\right].
\label{eq:shared-dynamics}
\end{equation}

\begin{theorem}[Spectral filtering under forgetting]
\label{thm:spectral-filtering}
Under Assumption~\ref{assump:shared-rpf}, let $\Sigma=\sum_i\lambda_i u_i u_i^\top$ be the eigendecomposition of this second-moment matrix, where $u_i$ are orthonormal shared pattern directions and $\lambda_i\ge0$ are their support strengths. Under Eq.~\eqref{eq:shared-dynamics}, the equilibrium is
\begin{equation}
w^*=\sum_i f(\lambda_i)\theta_i u_i,
\qquad
\theta_i=\theta^\top u_i,
\end{equation}
where
\begin{equation}
f(\lambda)=\frac{(1-d)\eta\lambda}{d+(1-d)\eta\lambda}.
\label{eq:spectral-filter}
\end{equation}
Thus there is a characteristic spectral threshold
\begin{equation}
\lambda^*=\Theta\!\left(\frac{d}{(1-d)\eta}\right)
\end{equation}
such that low-support modes satisfy $f(\lambda)=\mathcal{O}(\lambda)$ for $\lambda\ll\lambda^*$, while high-support modes satisfy $f(\lambda)\approx1$ for $\lambda\gg\lambda^*$.
\end{theorem}

Diagonalizing $\Sigma$ reduces each mode to the scalar recursion in Theorem~\ref{thm:exposure-selective} with $p(s)$ replaced by $\lambda_i$; see Appendix~\ref{app:proofs}. This lifts exposure-selective survival from isolated features to shared representations: forgetting filters the support spectrum, preserving reusable directions supported by recurring structures and attenuating weak or idiosyncratic modes.

\subsubsection{Information-constrained interpretation}

Third, under smooth multiplicative decay, the same dynamics also admits an information-constrained interpretation. Assume small $d$ and $\eta$, and standard Gaussian or coding approximations connecting parameter norm to description length. Under these assumptions, multiplicative forgetting,
\begin{equation}
W_{t+1}=(1-d)\left(W_t-\eta\nabla\mathcal{L}_t(W_t)\right).
\label{eq:wd-dynamics}
\end{equation}
to first order in the small-step limit follows gradient flow on an objective of the form
\begin{equation}
\mathcal{L}(W)+\frac{d}{2\eta}\|W\|^2.
\label{eq:l2-objective}
\end{equation}
Parameter norm can then proxy description length or stored information \citep{hinton1993keeping,rissanen1978modeling,grunwald2007minimum}. This is an approximate interpretation, not an exact equivalence to Information Bottleneck or MDL objectives \citep{tishby2015deep,alemi2017deep,kawaguchi2023information}. RPF provides a dynamics-based route to an MDL-like effect: information that is not repeatedly reconstructed is continuously leaked, whereas supported and recoverable information remains worth maintaining. The norm/coding assumptions and the corresponding approximation are detailed in Appendix~\ref{app:info}.

\section{Experiments}

The experiments test how repeated reinforcement and persistent forgetting jointly select what remains learned. We first examine exposure-selective survival in scalar memories, then conditional associations in a shared nonlinear predictor. Finally, matched reinforcement schedules test how the history of reinforcement changes the retained set at identical occurrence counts. The neural task has a uniform output marginal.

\subsection{Experiment 1: Independent-Feature Survival}

This experiment tests the basis-case mechanism in Section~3.3.1: whether nonuniform exposure, opposed by persistent attenuation, is sufficient to convert a heavy-tailed distribution over patterns into selective retention.

\paragraph{Setup and results.}
We simulate $N=10{,}000$ independent scalar memories with Zipf exposure $p(s_r)\propto r^{-1.2}$, target $\theta_s=1$, learning rate $\eta=0.2$, and batch-wise attenuation $w\leftarrow(1-d)w$ with $d=0.015$ every $32$ updates. For scaling runs, we vary the number of updates, retain the plotting coordinate $S=n\eta(1-d)$, and count a pattern as retained when $w_s>0.01\max_j w_j$. Appendix~\ref{app:exp-details} distinguishes this sampled, relative-threshold count from the theoretical cutoff on expected strength.

\begin{figure}[t]
\centering
\includegraphics[width=0.47\linewidth]{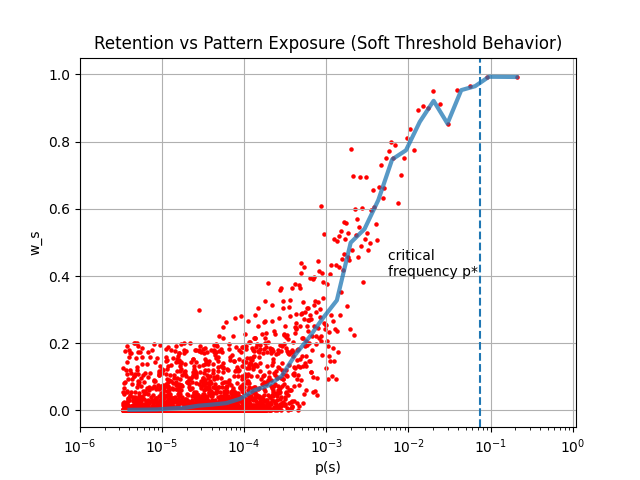}
\hfill
\includegraphics[width=0.47\linewidth]{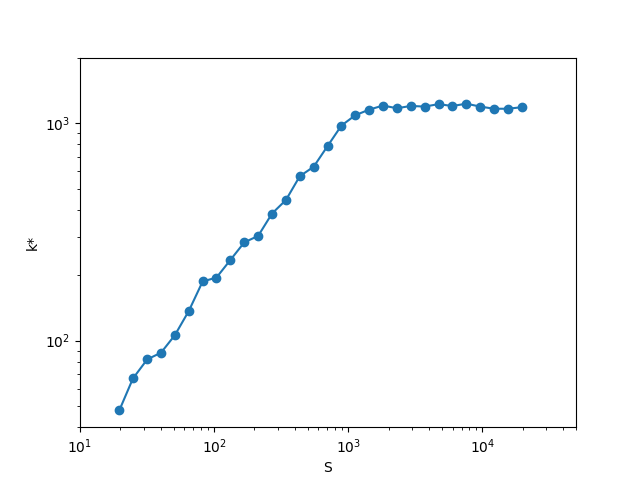}
\caption{Independent-feature signatures of RPF. Left: learned strength after $10^5$ updates exhibits the predicted soft support-dependent transition. Right: the number of sampled memories above a relative threshold grows sublinearly and approaches saturation, in qualitative agreement with the finite-time-to-steady-state picture. The expected-strength law $k^*\sim S^{1/\alpha}$ describes the accumulation-dominated regime. These batch-wise, finite-run observations support the predicted pattern of selective retention and growth toward a forgetting-limited ceiling.}
\label{fig:independent-results}
\end{figure}

In Figure~\ref{fig:independent-results}, each point in the left panel is one pattern and the smoothed curve averages logarithmic exposure bins. Low-exposure patterns receive too few reconstruction events to compensate for attenuation, whereas high-exposure patterns are repeatedly rebuilt and approach the target. The soft transition and intermediate-support scatter agree with the expected profile in Eq.~\eqref{eq:independent-fixed-point}. The right panel shows sublinear expansion and eventual saturation of the sampled retained set. This illustrates the RPF mechanism: reinforcement initially extends retention into the tail, while persistent forgetting sets the long-run support boundary. Section~\ref{sec:discussion} connects these two regimes analytically.

\subsection{Experiment 2: Conditional Retention under Reinforcement and Forgetting}

The independent-feature experiment isolates scalar survival, but real neural models share parameters across many patterns. We next ask whether support-dependent selection also appears for conditional associations in a shared nonlinear network. We construct the task so that the output marginal alone cannot predict the target. Each pattern $s=(r,k)$ combines one of $64$ Zipf ranks with one of $8$ labels, and its target is $y(s)=k$. Every rank has equal mass for each label, including after focal resampling. Thus $P(Y=k)=1/8$ and $H(Y\mid S)=0$: successful prediction requires the input--label relation. The network receives a single identifier for $s$, without direct access to $k$.

\paragraph{Setup and conditional control.}
A 24-dimensional embedding feeds a shared $24\!\to\!48\!\to\!8$ GELU head trained with cross-entropy. We designate ranks $3,7,\ldots,63$ as focal and cross their resampling weights $m\in\{1,2,4,8\}$ with seven AdamW decay coefficients over five seeds, for $140$ runs. Every run uses $600$ steps, batch size $256$, and learning rate $\eta=0.003$. The per-step parameter attenuation is
\begin{equation}
d_{\rm step}=\eta\lambda_{\rm wd}
\in\{0,0.003,0.006,0.009,0.012,0.015,0.021\}.
\label{eq:adamw-attenuation}
\end{equation}
We measure conditional strength $c_s=P_W(y(s)\mid s)$ and usable retention $c_s\ge0.5$. A same-rank control replaces $(r,k)$ with $(r,k+1\bmod8)$ while querying the original label $k$. Averaged over focal patterns across the full grid, correct-input probability is $0.694$, versus $0.043$ under this control, giving a conditional gap of $0.652$. This gap shows that the retention measure depends on learned input--label associations rather than output marginal fitting.

\begin{figure}[t]
\centering
\includegraphics[width=\linewidth]{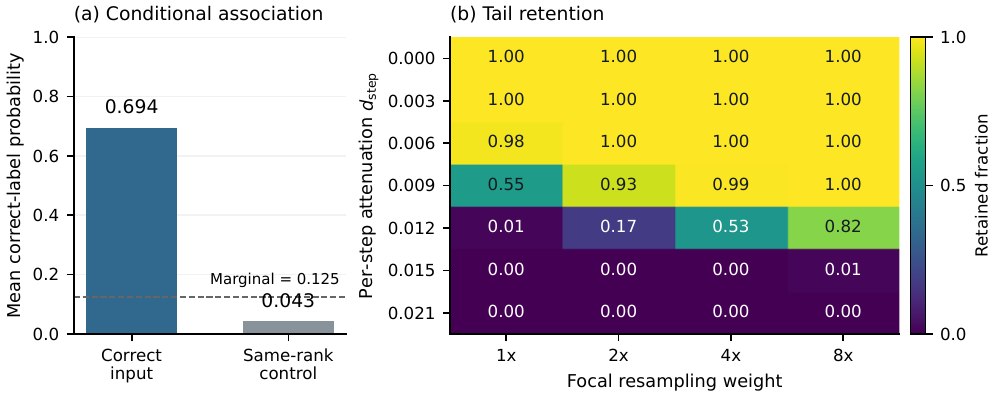}
\caption{Conditional retention and its reinforcement--forgetting phase diagram. Left: correct-input and same-rank control probabilities, averaged over focal patterns across all $140$ runs; the dashed line is the uniform label marginal. Right: tail focal retention ($c_s\ge0.5$), averaged over five seeds. Resampling moves associations into the retained region near the transition. The horizontal coordinate is the resampling weight, not the normalized exposure multiplier; the $8\times$ weight increases each focal pattern's sampling probability by $3.448\times$.}
\label{fig:conditional-phase}
\end{figure}

\paragraph{Selection boundary and intervention.}
Figure~\ref{fig:conditional-phase} exhibits a saturated low-attenuation region, a transition where reinforcement changes the retained set, and a high-attenuation region where the tested reinforcement is insufficient for tail retention. At $d_{\rm step}=0.012$, increasing $m$ from $1$ to $2$, $4$, and $8$ raises tail focal retention from $0.013$ to $0.175$, $0.525$, and $0.819$. On the continuous focal-probability scale, the $8\times-1\times$ gain is $0.014$ at zero attenuation, $0.177$ at $0.012$, and $0.041$ at $0.021$. The intervention is most effective near the retention transition. The observable RPF mapping is concrete: $s$ is a conditional association, $q_m(s)$ its normalized sampling probability, $c_s$ its observable strength, and $d_{\rm step}$ the controlled parameter attenuation; per-exposure recoverability $R(s)$ is not measured here. The same association can cross the retention criterion as its sampling support changes. Thus, in a network with shared parameters, attenuation shifts the support required for a conditional association to remain usable. Appendix~\ref{app:exp-details} gives all phase cells and sensitivity to thresholds $0.4$, $0.5$, and $0.6$. A factorized synthetic task also shows selective retention through held-out combinations without a complete-pattern embedding (Table~\ref{tab:factorized} and Appendix~\ref{app:exp-details}).

\subsection{Experiment 3: Temporal Coupling and Pattern Selection}

RPF also predicts that the final value of an occurrence depends on the dynamics that follow it. We therefore hold the complete sample multiset fixed and vary the temporal organization of reinforcement. For each of ten seeds, early, uniform, and late schedules share the same initialization, $600$ optimizer steps, and exact $153{,}600$-sample multiset, including $55{,}512$ additional focal replay samples. The remaining base samples also contain focal occurrences. The schedule changes the placement of the additional replay samples and hence local minibatch composition; every identifier has exactly the same count across schedules. Four attenuation levels yield $120$ runs.

Ordinary optimization order can matter even without decay. For mean focal conditional probability $Y$, we measure the additional attenuation-dependent timing difference by the paired contrast
\begin{equation}
\begin{aligned}
C(d_{\rm step})={}&
\big[Y_{\rm late}(d_{\rm step})-Y_{\rm early}(d_{\rm step})\big]
\\&-\big[Y_{\rm late}(0)-Y_{\rm early}(0)\big].
\end{aligned}
\label{eq:timing-contrast}
\end{equation}
The zero-attenuation late--early gap is $0.0186$. At $d_{\rm step}=0.009$, $0.012$, and $0.015$, the additional contrasts are respectively $0.1999$, $0.2895$, and $0.3009$, with paired 95\% bootstrap intervals $[0.1929,0.2069]$, $[0.2814,0.2978]$, and $[0.2934,0.3081]$. Equal occurrence counts therefore produce substantially different final retention, and the size of that difference depends on persistent forgetting.

\begin{figure}[t]
\centering
\includegraphics[width=\linewidth]{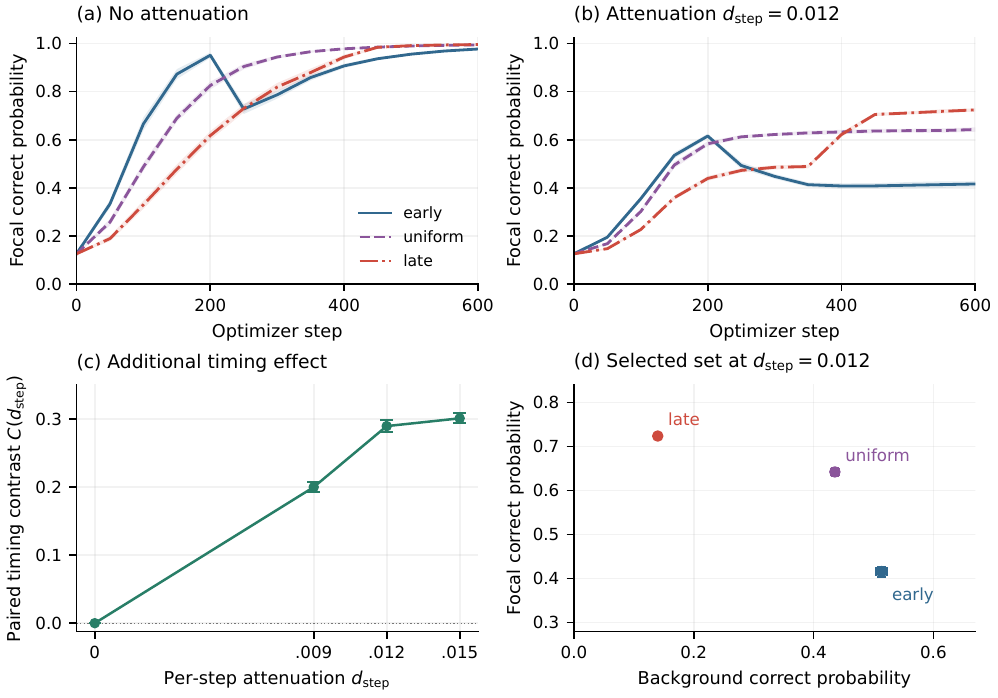}
\caption{Timing changes the retained value of identical sample multisets. Top: focal probability trajectories with zero attenuation and with $d_{\rm step}=0.012$; curves show ten-seed means and shading shows standard deviations. Bottom left: the paired timing contrast in Eq.~\eqref{eq:timing-contrast}, with 95\% bootstrap intervals across seeds. Bottom right: at $d_{\rm step}=0.012$, later focal replay increases focal correct-label probability while decreasing background correct-label probability; points are schedule means and bars are seed standard deviations. The intervention changes the composition of the retained set.}
\label{fig:temporal-selection}
\end{figure}

The trajectories in Figure~\ref{fig:temporal-selection} show the mechanism at the behavioral level: under attenuation, early focal gains decline during subsequent base training, while late reinforcement remains stronger at evaluation. The effect is selective. At $d_{\rm step}=0.012$, focal/background probabilities are $0.416/0.514$ for early, $0.642/0.436$ for uniform, and $0.724/0.140$ for late replay. Under the same $c_s\ge0.5$ criterion, mean focal/background retained fractions change from $0.232/0.465$ (early) to $0.995/0.000$ (late) across ten seeds. Rearranging the same training budget changes which associations remain strongly available. These results establish a forgetting-dependent temporal effect in a nonlinear shared network: equal occurrence counts do not guarantee equal final reinforcement or the same retained pattern set.

\section{Discussion}
\label{sec:discussion}

Our results characterize learning under forgetting as survival under the combined effects of training dynamics and the data distribution. RPF is therefore distinct from denoising or semantic filtering. It assumes no prior separation between signal and noise, and it does not assign value according to whether a stored item is meaningful, true, or useful. The survival criterion is support and recoverability: information remains encoded when it can be reconstructed often enough, and quickly enough, to offset persistent attenuation. High-support superficial patterns may survive, while rare but meaningful facts may decay unless they are replayed, retrieved, curated, or otherwise reinforced. Repeated artifacts or biases can also survive if they receive sufficient support. In this sense, RPF describes a passive selection effect induced by training dynamics, not a criterion for truth, utility, or semantic importance. Here passive selection means that the dynamics determine which patterns lose recoverability without explicitly prescribing which individual patterns must be forgotten. Adjusting a global attenuation hyperparameter changes the selection pressure; the identities of lost patterns emerge from its interaction with reinforcement.

Under RPF, abstraction corresponds to repeated reconstruction of shared directions under attenuation. RPF does not define abstraction semantically or impose it architecturally; abstraction appears as the residue of repeated reconstruction under forgetting. In shared parameters, the unit of survival changes from isolated patterns to covariance-supported directions: reusable directions persist, while weak or idiosyncratic directions are attenuated. In the factorized synthetic task, added reinforcement also increases correct-label probability through held-out template combinations without a complete-association embedding (Appendix~\ref{app:exp-details}).

RPF is related to regularization, but it is not merely a restatement of weight decay. The novelty is that attenuation coupled to nonuniform reinforcement converts uniform forgetting pressure into a nonuniform survival boundary. In the regularization view, weight decay is usually a small penalty for generalization; in the RPF view, it is a controllable forgetting pressure that can be increased to raise the support required for retention.

\paragraph{Reinforcement value depends on history.}
In RPF, repeated reinforcement continually reconstructs information while persistent forgetting makes reconstruction necessary for continued retention. The strength and temporal organization of compatible updates determine their effective reinforcing contribution under this forgetting pressure. In the scalar case, write $\mathcal U^{(a)}(w)=(1-a)w+a\theta$ and $\mathcal F^{(d)}(w)=(1-d)w$. Then
\begin{equation}
\mathcal F^{(d)}(\mathcal U^{(a)}(w))-
\mathcal U^{(a)}(\mathcal F^{(d)}(w))=-ad\theta.
\label{eq:commutator}
\end{equation}
For a sequence of exposures, the final state weights each update by the dynamics that follow it (Appendix~\ref{app:theory-details}). Equal occurrence counts can therefore supply different effective reinforcement under persistent forgetting. The timing result refines repeated reinforcement itself: its effective value is the part of compatible updates that survives subsequent dynamics. Experiment~2 tests this reconstruction--attenuation balance through joint interventions; Experiment~3 tests the temporal organization of reinforcement within the same coupling. The resulting changes in conditional retention and retained-set composition provide controlled empirical evidence for the RPF selection framework.

\paragraph{Finite-time expansion and the steady-state ceiling.}
The retained-count law and the stationary boundary describe different regimes of the same recursion. For $w_s^0=0$, $\theta_s\ne0$, and $\beta_s=(1-d)\eta p(s)$, the exact normalized expected strength is
\begin{equation}
r_n(s):=\frac{\mathbb E[w_s^n]}{\theta_s}
=\frac{\beta_s}{d+\beta_s}
\left[1-(1-d-\beta_s)^n\right].
\label{eq:finite-exact}
\end{equation}
For $nd\ll1$ and small learning steps near the boundary, $r_n(s)\approx1-\exp[-n\beta_s]$. Thus a fixed criterion $r_n(s)\ge\epsilon$ gives $p_\epsilon(n)\approx-\log(1-\epsilon)/S$ and the accumulation-dominated law $k^*\sim S^{1/\alpha}$. At fixed positive $d$, however, $r_n$ approaches $\beta_s/(d+\beta_s)$, and the boundary approaches $p_\epsilon^*$ in Eq.~\eqref{eq:epsilon-boundary}. The limiting retained rank is of order $(C/p_\epsilon^*)^{1/\alpha}$ within the available rank range. Further training approaches this forgetting-limited ceiling. Experiment~1 uses batch-wise attenuation and a relative threshold on sampled strengths. Its observed sublinear expansion and saturation visualize this RPF progression: reinforcement initially expands the retained set, while persistent forgetting sets its long-run support boundary.

\paragraph{Recoverability conditional on exposure.}
The factors in $q(s)R(s)$ must count exposure only once. At a fixed local state and update protocol, our convention is
\begin{equation}
\mathbb E[\Delta w_s^{\mathcal U,\mathrm{direct}}\mid E_t(s)=e,Z_t]
\approx\eta e R(s;Z_t)(\theta_s-w_s^t),
\label{eq:conditional-recovery}
\end{equation}
where $e$ is the minibatch exposure fraction, $\Delta w_s^{\mathcal U,\mathrm{direct}}$ is the compatible contribution before the forgetting operator, and $\theta_s$ is a specified reference strength. We suppress the state argument in the mean-field notation. Here $R$ measures normalized recovery per exposure; averaging the direct contribution over minibatches gives the scale $\eta qR$. Recoverability reflects the architecture, optimizer, current representation, context diversity, and the extent to which other patterns share the same structure. The independent-feature case has $R=1$. In contrast, $1/T_{\rm rec}^{\rm stream}$, with time measured in ordinary training steps, is a process recovery rate that already includes exposure waiting, intervening attenuation, and shared updates. It cannot be substituted for $R$ and multiplied by $q$ again. A specified probe and local update assay give the corresponding operational definition in a nonlinear network.

\paragraph{Empirical support and quantitative questions.}
The conditional neural experiments provide evidence for the RPF framework: compatible reinforcement and persistent forgetting jointly select which input--label associations remain available under a shared nonlinear head. The phase diagram shows how reinforcement and attenuation move conditional associations across a retention transition; matched schedules show that the same exposure count can select a different retained set. At the quantitative level, the nonlinear phase boundary moves toward higher support as attenuation increases, with an empirical log--log slope of $3.05$ across $53$ runs with identifiable crossings, compared with the approximately unit slope of the small-decay scalar boundary at fixed learning rate. This is a boundary-versus-attenuation slope, distinct from the count-versus-training-strength exponent $1/\alpha$. The fixed ratio $\rho=(1-d_{\rm step})\eta q/d_{\rm step}$ also fails to collapse the AdamW curves; its held-out-seed probability RMSE is $0.1767$, compared with $0.1072$ for a decay-only predictor. These results support the RPF selection picture and identify the quantitative role of nonlinear training dynamics as a further question.

The neural measurements concern conditional behavior. Identifying $R$ and local support operators, testing the covariance-mode law, and evaluating generalization across broader compositions and real language data are the next empirical steps. The paired timing contrast measures the schedule-by-attenuation interaction, including effects mediated by adaptive optimization and shared-gradient interference: the replay blocks change the local minibatch distribution even though occurrence counts match. Separating the contributions of these pathways requires additional interventions. AdamW controls parameter attenuation; its effect on conditional probability is mediated by the network and optimizer.

The heavy-tailed analysis should also be read mechanistically. The exact Zipf or power-law form is not required for RPF itself; it only yields a closed-form retained-count scaling. The more general requirement is scale separation in effective support, which may arise from pattern exposure, phrase and relation reuse, visual or acoustic motifs, contextual diversity, replay, weighting, or covariance-supported representation directions. Under heavy-tailed exposure, a small retained type set can still cover a large fraction of exposure mass.

\paragraph{Implications for controlling retention.}
RPF treats attenuation as a controllable forgetting pressure and replay as a way to allocate reinforcement across patterns. Additional training can expand the retained set until the forgetting-limited ceiling; sampling and weighting can increase support for selected patterns, while stronger attenuation raises the support needed to retain them. The phase diagram identifies where added support has the greatest effect, and the timing experiment shows that the final value of a fixed reinforcement budget depends on when it is spent. Concentrated late reinforcement can preserve selected associations at the expense of background retention, so the retained set must be evaluated alongside focal gains. Norm/coding approximations further motivate an MDL-like interpretation: repeated reinforcement maintains supported information despite the cost of storing it, while insufficiently reconstructed components decay. The central result remains the dynamically selected retained set.

\section{Conclusion}

RPF formulates learning as repeated reconstruction under persistent forgetting. The independent-feature model gives an exposure-dependent survival boundary, while the shared-linear model shows how this selection preserves covariance-supported directions rebuilt by recurring structures. Finite-time dynamics connect initial retained-set expansion to a forgetting-limited ceiling. Conditional neural experiments establish a reinforcement--forgetting phase diagram and show that identical occurrence counts produce different retention when the temporal organization of reinforcement changes. The same intervention changes which patterns remain available, providing empirical support for the framework. RPF therefore identifies a concrete selection principle: attenuation sets the support required for retention, while the amount and timing of reinforcement determine which patterns cross that boundary.

\appendix

\section{Proofs}
\label{app:proofs}

\paragraph{Proof of Theorem~\ref{thm:exposure-selective}.}
Taking conditional expectation in Eq.~\eqref{eq:independent-dynamics} gives
\begin{align}
\mathbb{E}[w_s^{t+1}\mid w_s^t]
&=(1-d)\left[w_s^t+\eta p(s)(\theta_s-w_s^t)\right]\\
&=(1-d)(1-\eta p(s))w_s^t+(1-d)\eta p(s)\theta_s.
\end{align}
Let $\mu_t=\mathbb{E}[w_s^t]$. Then
\begin{equation}
\mu_{t+1}=a_s\mu_t+b_s,
\quad
a_s=(1-d)(1-\eta p(s)),\quad b_s=(1-d)\eta p(s)\theta_s.
\end{equation}
Since $0<a_s<1$, the recursion converges to $\mu^*=b_s/(1-a_s)$, yielding Eq.~\eqref{eq:independent-fixed-point}. The threshold follows by comparing $(1-d)\eta p(s)$ with $d$.

\paragraph{Proof of Theorem~\ref{thm:spectral-filtering}.}
Project Eq.~\eqref{eq:shared-dynamics} onto an eigenvector $u_i$ and write $w_i^t=(w^t)^\top u_i$. Since $\Sigma u_i=\lambda_i u_i$,
\begin{equation}
w_i^{t+1}=(1-d)\left[(1-\eta\lambda_i)w_i^t+\eta\lambda_i\theta_i\right].
\end{equation}
This is the same scalar recursion as in Theorem~\ref{thm:exposure-selective} with $p(s)$ replaced by $\lambda_i$. Solving the fixed point yields Eq.~\eqref{eq:spectral-filter}.

\section{Additional Theoretical Details}
\label{app:theory-details}

\paragraph{Pattern scope and forgetting operators.}
The pattern unit $s$ used in Section~3 is operational: its occurrences induce compatible learning signals. Discrete identifiers and scalar features are controllable basis cases; covariance-supported directions provide the corresponding object for the shared-linear analysis. The operator formulation in Eq.~\eqref{eq:rpf-general} accommodates different implementations of persistent forgetting. An independent Bernoulli erasure of stored state with probability $d$ has conditional mean attenuation $1-d$, so multiplicative decay gives a tractable mean-field instance of this broader formulation. Each implementation determines its finite-run variability and its mapping from stored state to behavioral forgetting.

\paragraph{Exact finite-time solution and retained-count regimes.}
For zero initialization and $\theta_s\ne0$, the expected scalar recurrence is
\begin{equation}
r_{t+1}=a_s r_t+\beta_s,
\qquad a_s=1-d-\beta_s,
\qquad \beta_s=(1-d)\eta p(s).
\end{equation}
Summing the geometric series gives Eq.~\eqref{eq:finite-exact}. Under the stated assumption $0<\eta p(s)<1$, $0<a_s<1$, so expected normalized strength increases toward $r_\infty=\beta_s/(d+\beta_s)$. The finite-horizon support boundary is defined implicitly by $r_n(p_\epsilon(n))=\epsilon$ wherever that level is attainable within the available support range. It approaches the steady-state boundary in Eq.~\eqref{eq:epsilon-boundary}.

In the accumulation-dominated regime, $nd\ll1$ and a small learning step at the boundary give
\begin{equation}
r_n(p)\approx1-e^{-n(1-d)\eta p},
\qquad
p_\epsilon(n)\approx\frac{-\log(1-\epsilon)}{n(1-d)\eta}.
\end{equation}
For $p(s_r)=C r^{-\alpha}$, the expected-strength cutoff is consequently
\begin{equation}
k^*(n)\approx
\left(\frac{CS}{-\log(1-\epsilon)}\right)^{1/\alpha},
\qquad S=n(1-d)\eta,
\end{equation}
while these conditions hold and the cutoff lies within the ranked support. At fixed $d>0$, the limiting rank boundary is instead $(C/p_\epsilon^*)^{1/\alpha}$, clipped to the available ranks. If even the largest available exposure falls below $p_\epsilon^*$, no pattern meets the stationary criterion; a pattern exactly at the boundary approaches the criterion asymptotically. This distinction separates finite-time growth from equilibrium selection.

The retained exposure mass satisfies
\begin{equation}
\sum_{r=1}^{k^*}p(s_r)=1-\mathcal O\big((k^*)^{1-\alpha}\big).
\label{eq:retained-mass}
\end{equation}
For a sufficiently large power-law support, substituting the early-regime cutoff yields $1-\mathcal O(S^{-(\alpha-1)/\alpha})$ in that regime. At fixed positive forgetting, the retained mass also approaches a ceiling determined by the stationary retained set.

\paragraph{Batch-wise attenuation.}
Experiment~1 applies forgetting every $B=32$ independent exposure events. Let $m_j$ be expected strength immediately after the $j$th attenuation. With $h_s=(1-\eta p(s))^B$, its exact cycle-level recurrence is
\begin{equation}
m_{j+1}=(1-d)\big[h_s m_j+(1-h_s)\theta_s\big].
\end{equation}
Hence, for zero initialization,
\begin{equation}
\frac{m_j}{\theta_s}
=\frac{(1-d)(1-h_s)}{1-(1-d)h_s}
\left[1-\big((1-d)h_s\big)^j\right].
\label{eq:batch-finite}
\end{equation}
For a final partial cycle of $\ell<B$ updates, apply the unattenuated expected update $\ell$ times to $m_j$. Matching attenuation alone gives the equivalent per-update rate $d_{\rm eff}=1-(1-d)^{1/B}\approx d/B$ for small $d$; the full cycle-level dynamics also include the intervening reinforcement through $h_s$. The plotted relative threshold counts realized sample strengths relative to their maximum, while Eqs.~\eqref{eq:finite-exact} and \eqref{eq:batch-finite} characterize the expected-strength profile. The observed soft transition, sublinear expansion, and saturation provide qualitative support for this reinforcement--forgetting picture.

The scatter in the independent-feature simulation also has a reconstruction-based interpretation. Low-support patterns receive few updates, while highly supported patterns approach saturation and have small recovery gaps. Around the observed transition, intermittent exposures combine with appreciable recovery gaps, producing visible variation in learned strength. This explains the intermediate-support scatter in Figure~\ref{fig:independent-results} at the level of the simulated mechanism.

\paragraph{Reinforcement weighted by subsequent dynamics.}
Let $I_t\in\{0,1\}$ be the exposure indicator for one scalar pattern, with time indexed by $t=0,\ldots,n-1$. Set
\begin{equation}
A_t=(1-d)(1-\eta I_t),\qquad c_t=(1-d)\eta I_t\theta.
\end{equation}
The realized update is $w_{t+1}=A_t w_t+c_t$. Repeated substitution gives the exact identity
\begin{equation}
w_n=\left(\prod_{u=0}^{n-1}A_u\right)w_0
+\sum_{t=0}^{n-1}c_t\prod_{u=t+1}^{n-1}A_u.
\label{eq:time-weighted}
\end{equation}
Empty products equal one. For $0<d<1$ and $0<\eta\le1$, each $A_t\in[0,1)$ is a contraction, and the final contribution of an exposure explicitly includes all subsequent contractions. Define
\begin{equation}
\widetilde\rho_n=\sum_{t=0}^{n-1}I_t\prod_{u=t+1}^{n-1}A_u.
\end{equation}
Then the accumulated input contribution is $(1-d)\eta\theta\widetilde\rho_n$. If subsequent reinforcement contractions are negligible, the weights reduce approximately to $(1-d)^{n-1-t}$. The exact expression also accounts for contraction from later reinforcement. Equal $\sum_t I_t$ therefore need not imply equal $w_n$ under forgetting. With zero decay and a single compatible scalar target, the complete scalar result depends only on the number of exposures; nonlinear shared optimization can retain order effects even at zero decay, motivating the experimental baseline subtraction.

Equation~\eqref{eq:commutator} follows by evaluating the two affine operators: reinforcement followed by forgetting has constant term $(1-d)a\theta$, whereas forgetting followed by reinforcement has constant term $a\theta$. Their difference is $-ad\theta$. These are consequences of the analyzed scalar dynamics and connect its stationary boundary to the finite-time timing prediction.

\paragraph{Shared linear dynamics and local nonlinear interpretation.}
For a possibly time-dependent affine shared update $w_{t+1}=A_t w_t+c_t$, define $\Phi(n,t+1)=A_{n-1}\cdots A_{t+1}$ and $\Phi(n,0)=A_{n-1}\cdots A_0$. The exact expansion is
\begin{equation}
w_n=\Phi(n,0)w_0+\sum_{t=0}^{n-1}\Phi(n,t+1)c_t.
\end{equation}
For the squared-loss shared-linear case, $A_t=(1-d)(I-\eta\Sigma_t)$. A linear pattern readout $v_s^\top w_n$ assigns update $c_t$ the final contribution $v_s^\top\Phi(n,t+1)c_t$, which depends on alignment as well as subsequent attenuation and shared updates. A local quadratic or Gauss--Newton approximation motivates analogous operators for a nonlinear network, with the reference-point terms included in $c_t$. Estimating those operators is a further identification task. The neural experiments test the observable conditional-selection and timing predictions without assuming a fixed nonlinear covariance spectrum.

\section{Information-Constrained Interpretation}
\label{app:info}

Expanding Eq.~\eqref{eq:wd-dynamics} gives
\begin{equation}
W_{t+1}-W_t=-\eta\nabla\mathcal{L}_t(W_t)-dW_t
+d\eta\nabla\mathcal{L}_t(W_t).
\end{equation}
For small $d$ and $\eta$, the last term is higher order. When the stochastic gradient represents the gradient of $\mathcal{L}$, the remaining terms are an Euler step of size $\eta$ for gradient flow on $\mathcal{L}(W)+\frac{d}{2\eta}\|W\|^2$. Under a fixed Gaussian coding approximation, the quadratic norm term can proxy a cost of storing parameter information. Repeated, loss-reducing updates must offset that cost for a component to persist; unsupported components decay. This is the dynamics-based fitting--storage trade-off behind the MDL-like interpretation, not an exact equivalence to IB or MDL objectives.

\section{Experimental Details}
\label{app:exp-details}

\paragraph{Procedure for Experiment~1.}
The independent-feature simulation constructs a normalized Zipf exposure vector over $N=10{,}000$ patterns,
\begin{equation}
p(s_r)=\frac{r^{-1.2}}{\sum_{j=1}^{N}j^{-1.2}}.
\end{equation}
All memories start at $w_s=0$ with targets $\theta_s=1$. At each update, sample $s\sim p$ and reinforce the sampled coordinate by $w_s\leftarrow w_s+\eta(1-w_s)$ with $\eta=0.2$. Every $32$ updates, attenuate all memories by $w\leftarrow(1-0.015)w$. The retained count uses $w_s>0.01\max_jw_j$. The original scaling plot sweeps $30$ logarithmically spaced update counts from $10^2$ to $10^5$. Its horizontal coordinate $S=n\eta(1-d)$ retains the original batch-level $d$ as a plotting convention; the cycle-level expectation is Eq.~\eqref{eq:batch-finite}. At the largest plotted horizon, cumulative forgetting is substantial, so the full curve cannot be treated as a test confined to the accumulation-dominated regime.

\paragraph{Conditional task and normalized support in Experiment~2.}
For $r\in\{1,\ldots,64\}$ and $k\in\{0,\ldots,7\}$, define $s=(r,k)$ and $y(s)=k$. A single integer identifier encodes $s$. Its base support is
\begin{equation}
p(r,k)=\frac{r^{-1.15}}{8\sum_{j=1}^{64}j^{-1.15}}.
\end{equation}
The focal ranks are $3,7,\ldots,63$, giving $128$ focal and $384$ background patterns. The tail focal set contains $32$ patterns at ranks $51,55,59,63$, within the bottom quarter of the rank range. Let $\mathcal F$ denote the focal set and $P_{\mathcal F}=\sum_{s\in\mathcal F}p(s)=0.1886452892$. With focal resampling weight $m$, the sampling distribution is
\begin{equation}
q_m(s)=\frac{p(s)m^{\mathbf1\{s\in\mathcal F\}}}
{1+(m-1)P_{\mathcal F}}.
\label{eq:normalized-support}
\end{equation}
For focal patterns, the true exposure multipliers for $m=1,2,4,8$ are $1$, $1.6826$, $2.5544$, and $3.4475$. Background sampling probability decreases through the same normalization. Because each rank contains every label with equal mass, both $p$ and $q_m$ have exactly uniform label marginals. Finite sampled label counts fluctuate around this distribution. The same-rank control changes the input label identity while preserving its sampling support, and queries the original label.

\paragraph{Network and optimization.}
The model consists of a $512\times24$ embedding table and a shared linear layer $24\to48$, GELU activation, and linear layer $48\to8$, with biases. It has $13{,}880$ trainable parameters and uses no dropout. All parameters are passed to AdamW with learning rate $0.003$, default moment coefficients $(0.9,0.999)$ and numerical constant $10^{-8}$. Each run minimizes mean minibatch cross-entropy for $600$ steps of batch size $256$. The grid is $\lambda_{\rm wd}\in\{0,1,2,3,4,5,7\}$, $m\in\{1,2,4,8\}$, and seeds $0$--$4$. Seeds match initialization across conditions; sampling generators depend on seed, resampling weight, and decay coefficient, so the phase-grid sample sequences are not matched. Experiment~3 supplies the exactly matched sequence intervention.

For AdamW, the parameter update has the form
\begin{equation}
W_{t+1}=(1-d_{\rm step})W_t-\eta g_t^{\rm Adam},
\end{equation}
where $g_t^{\rm Adam}$ is the adaptive update direction. This controls parameter attenuation directly. Holding the update direction fixed, post-update attenuation in Eq.~\eqref{eq:wd-dynamics} would also scale the update by $1-d_{\rm step}$; the one-step difference is $d_{\rm step}\eta g_t^{\rm Adam}$. Thus the scalar operator identity is exact in its basis case, while the neural timing experiment tests the corresponding selection effect under AdamW without imposing the scalar coefficients on conditional probabilities.

\paragraph{Metrics and full phase grid.}
All probabilities are evaluated with the model in evaluation mode. Primary strength is $c_s=P_W(y(s)\mid s)$, and usable retention is $\mathbf1\{c_s\ge0.5\}$. Group metrics are unweighted means across identifiers within the focal, tail focal, or background group, followed by means across seeds. Table~\ref{tab:phase} gives every tail phase cell, with sample standard deviations across the five seeds. The pooled correct-input and control means in Figure~\ref{fig:conditional-phase} are over $128\times140=17{,}920$ focal observations.

\begin{table}[htbp]
\centering
\small
\caption{Complete tail focal retention grid, mean $\pm$ seed standard deviation. The retention criterion is $c_s\ge0.5$; columns are focal resampling weights.}
\label{tab:phase}
\begin{tabular}{rcccc}
\toprule
$d_{\rm step}$ & $1\times$ & $2\times$ & $4\times$ & $8\times$ \\
\midrule
0.000 & $1.000\pm0.000$ & $1.000\pm0.000$ & $1.000\pm0.000$ & $1.000\pm0.000$ \\
0.003 & $1.000\pm0.000$ & $1.000\pm0.000$ & $1.000\pm0.000$ & $1.000\pm0.000$ \\
0.006 & $0.981\pm0.028$ & $1.000\pm0.000$ & $1.000\pm0.000$ & $1.000\pm0.000$ \\
0.009 & $0.550\pm0.084$ & $0.925\pm0.036$ & $0.994\pm0.014$ & $1.000\pm0.000$ \\
0.012 & $0.013\pm0.017$ & $0.175\pm0.052$ & $0.525\pm0.124$ & $0.819\pm0.056$ \\
0.015 & $0.000\pm0.000$ & $0.000\pm0.000$ & $0.000\pm0.000$ & $0.013\pm0.028$ \\
0.021 & $0.000\pm0.000$ & $0.000\pm0.000$ & $0.000\pm0.000$ & $0.000\pm0.000$ \\
\bottomrule
\end{tabular}

\end{table}

\begin{table}[htbp]
\centering
\small
\caption{Threshold sensitivity at $d_{\rm step}=0.012$. Tail focal retention means across five seeds. The intervention direction persists across all three criteria.}
\label{tab:threshold}
\begin{tabular}{rcccc}
\toprule
Criterion & $1\times$ & $2\times$ & $4\times$ & $8\times$ \\
\midrule
0.4 & 0.200 & 0.706 & 0.969 & 1.000 \\
0.5 & 0.013 & 0.175 & 0.525 & 0.819 \\
0.6 & 0.000 & 0.013 & 0.069 & 0.125 \\
\bottomrule
\end{tabular}

\end{table}

Table~\ref{tab:threshold} checks criteria $0.4$, $0.5$, and $0.6$. The continuous $m=8$ minus $m=1$ focal probability gain provides a threshold-free counterpart: $0.014$ at zero attenuation, $0.177$ at $0.012$, and $0.041$ at $0.021$.

\paragraph{Empirical boundary and ratio analysis.}
For each run containing both retained and non-retained focal identifiers, we fit a logistic classifier of retention on $\log q_m(s)$, with intercept, inverse regularization strength $10^4$, and maximum $2{,}000$ iterations. Its 0.5 fitted-probability crossing defines $p_{50}$. A log--log regression across the $53$ eligible runs gives slope $3.047$ and $R^2=0.871$ for boundary versus decay coefficient; using $d_{\rm step}$ changes the intercept but not the slope because learning rate is fixed. Runs with no crossing are not assigned a boundary. This slope describes the empirical transition under nonlinear AdamW training and does not estimate the retained-count exponent $1/\alpha$.

A post-hoc predictor comparison used leave-one-seed-out evaluation of continuous focal probabilities at the six nonzero-decay levels, holding out all observations of each seed together. All five models use the same observations. Cubic spline features with six knots and a ridge penalty of $10^{-3}$ describe log support, log attenuation, their additive or interaction combinations, or the fixed log-ratio coordinate. The mean held-out RMSEs are $0.2591$ for support alone, $0.1072$ for decay alone, $0.1767$ for the fixed ratio coordinate, $0.0612$ for a flexible two-axis additive model, and $0.0405$ for a flexible two-axis interaction model. These exploratory fits diagnose limits to quantitative transfer rather than define coupling: statistical interaction depends on the response scale, whereas Experiment~3 intervenes on the timing of matched reinforcement events.

\paragraph{Matched schedules in Experiment~3.}
The network, learning rate, batch size, and number of steps match Experiment~2. We use $\lambda_{\rm wd}\in\{0,3,4,5\}$, schedules early/uniform/late, and seeds $0$--$9$. For each seed, a base pool is sampled from $p$ and an extra replay pool from $p(s\mid s\in\mathcal F)$. For nominal weight $m=4$, mixing fraction
\begin{equation}
\mu=\frac{(m-1)P_{\mathcal F}}{1+(m-1)P_{\mathcal F}}
=0.3614042435
\end{equation}
reproduces $q_4$ in expectation. Rounding $\mu\times153{,}600$ gives $55{,}512$ additional replay samples and $98{,}088$ base samples. The base pool itself includes focal patterns. We place the replay pool first, approximately evenly interleaved, or last, while preserving the internal order of each pool. Exact per-identifier counts are identical across schedules for every seed, and all decay levels reuse these same sequences. Each run resets the model and optimizer to the same seed-specific initialization and fresh optimizer state. The design was chosen using the completed phase grid; the schedule experiment then used the fixed configuration for all ten seeds.

\paragraph{Paired inference and selection.}
For each seed and decay, subtract that seed's zero-decay late--early gap from its corresponding nonzero-decay gap. Confidence intervals resample the ten paired seed contrasts with replacement $5{,}000$ times and use the 2.5th and 97.5th percentiles of the bootstrap means. The bootstrap generator seed is $20260729$. Seeds, rather than individual identifiers, are the independent resampling unit. Table~\ref{tab:timing} gives all primary contrasts. Trajectories are recorded at initialization and every $50$ optimizer steps. The focal/background tradeoff in Figure~\ref{fig:temporal-selection} uses final group means at $d_{\rm step}=0.012$.

\begin{table}[H]
\centering
\small
\caption{Matched schedule effects on mean focal conditional probability. The contrast subtracts each seed's zero-attenuation late--early gap. Intervals are paired-seed bootstrap intervals.}
\label{tab:timing}
\begin{tabular}{rrrc}
\toprule
$d_{\rm step}$ & Late--early & $C(d_{\rm step})$ & 95\% interval \\
\midrule
0.000 & 0.0186 & 0.0000 & --- \\
0.009 & 0.2185 & 0.1999 & $[0.1929,\,0.2069]$ \\
0.012 & 0.3081 & 0.2895 & $[0.2814,\,0.2978]$ \\
0.015 & 0.3195 & 0.3009 & $[0.2934,\,0.3081]$ \\
\bottomrule
\end{tabular}

\end{table}

\paragraph{Factorized held-out combinations.}
In a separate synthetic task, each input combines one of $64$ Zipf-ranked entities, one of $8$ relations, and one of $4$ shared templates. Its target is $y=(r\bmod8+\mathrm{relation})\bmod8$, invariant across templates. Each entity--relation association is trained through three templates and evaluated through the fourth: all factors appear in training, but none of the $512$ held-out triples does. Separate entity, relation, and template embeddings feed a shared GELU network, with no complete-association embedding. Training and held-out label marginals are uniform. We cross resampling weights $m\in\{1,4\}$ with AdamW decay coefficients $\{0,3,4,5\}$ over ten seeds, giving $80$ runs.

Within each seed and resampling condition, the initialization and sampled sequence are matched across decay levels. Table~\ref{tab:factorized} reports paired gains in focal correct-label probability on held-out combinations. The gain peaks near the decay-$3$ transition at $0.374$; the corresponding held-out background gain is $0.029$ with interval $[-0.009,0.070]$. This extends the conditional selection observation to associations reconstructed from shared factors through unseen entity--relation--template combinations; each template itself appears in training.

\begin{table}[htbp]
\centering
\small
\caption{Factorized synthetic task: paired gain in focal correct-label probability on held-out combinations for resampling weight $m=4$ versus $m=1$ (ten seeds; paired 95\% bootstrap intervals).}
\label{tab:factorized}
\begin{tabular}{ccc}
\toprule
AdamW decay coefficient & Focal gain & 95\% interval \\
\midrule
$0$ & $0.165$ & $[0.136,0.199]$ \\
$3$ & $0.374$ & $[0.267,0.479]$ \\
$4$ & $0.158$ & $[0.094,0.222]$ \\
$5$ & $0.055$ & $[0.052,0.058]$ \\
\bottomrule
\end{tabular}
\end{table}

\begin{samepage}
\paragraph{Compute and reproducibility.}
The recorded neural runs used a CPU on macOS 14.5 (arm64), Python 3.9.18, PyTorch 2.3.0, NumPy 2.0.2, pandas 2.2.3, scikit-learn 1.6.1, and Matplotlib 3.9.4, with four PyTorch threads. The $140$ phase runs and $120$ schedule runs comprise $156{,}000$ optimizer steps and $39{,}936{,}000$ sampled training presentations; the factorized study adds $80$ runs of $600$ steps each. The phase and timing figures and tables are regenerated from saved run-level and identifier-level outputs, and Table~\ref{tab:factorized} is checked against the saved paired summary. The revision checks matched counts, reported contrasts, threshold summaries, and the exact recursion identities. These checks verify consistency of the saved evidence and analytical expressions; they do not constitute a new execution of the training runs.
\par\end{samepage}

\end{document}